\documentclass[10pt,journal,compsoc,twoside]{IEEEtran}
\ifCLASSOPTIONcompsoc
\usepackage[nocompress]{cite}
\else
\usepackage{cite}
\fi
\ifCLASSINFOpdf
\else
\fi
\usepackage {booktabs}
\usepackage {graphicx}
\usepackage{subfigure}
\usepackage{latexsym}
\usepackage{setspace}
\usepackage{subfigure}
\usepackage{amsmath,amsfonts,amscd,amssymb}
\usepackage[noend]{alg-clipse}
\usepackage{alg-clipse}
\usepackage{multirow}
\usepackage{algorithm}
\usepackage{chapterbib}
\usepackage[none]{hyphenat}
\usepackage{url}
\usepackage{array}

\usepackage{chapterbib}
\usepackage{epstopdf}
\usepackage{caption}
\usepackage{setspace}
\usepackage{verbatim}
\usepackage{float}

\renewcommand{\algorithmicrequire}{\textbf{Input:}} 
\renewcommand{\algorithmicensure}{\textbf{Output:}} 
\usepackage{ragged2e}

\usepackage{geometry}
\usepackage{diagbox}
\usepackage{multirow}
\newtheorem{definition}{\hspace{0em}Definition}

\begin{document}
\title{GISExplainer: On Explainability of Graph Neural Networks via Game-theoretic Interaction Subgraphs}

\author{Xingping~Xian,~\IEEEmembership{}
Jianlu~Liu,~\IEEEmembership{}
Chao~Wang,~\IEEEmembership{}
Tao~Wu,~\IEEEmembership{}
Shaojie~Qiao,~\IEEEmembership{}
Xiaochuan~Tang,~\IEEEmembership{}
Qun~Liu~\IEEEmembership{}

\IEEEcompsocitemizethanks{

\IEEEcompsocthanksitem X. Xian, J. Liu, and T. Wu are with School of Cyber Security and Information Law, Chongqing University of Posts and Telecommunications, Chongqing, China. (E-mail: wutaoadeny@gmail.com, xxp0213@gmail.com).

\IEEEcompsocthanksitem Q. Liu is with School of Computer Science and Technology, Chongqing University of Posts and Telecommunications, Chongqing, China.

\IEEEcompsocthanksitem X. Tang is with Chengdu University of Technology, Chengdu, China.

\IEEEcompsocthanksitem S. Qiao is with School of Software Engineering, Chengdu University of Information Technology, Chengdu, China. E-mail: qiaoshaojie@gmail.com.

\IEEEcompsocthanksitem C. Wang is with School of Computer and Information Science, Chongqing Normal University, Chongqing, China. E-mail: chaosimpler@gmail.com.



\IEEEcompsocthanksitem This work was supported in part by the National Natural Science Foundation of China (No.62106030, 62376047); Key Projects of Chongqing Natural Science Foundation Innovation and Development Joint Fund (No.CSTB2023NSCQ-LZX0003); Key Project of Science and Technology Research Program of Chongqing Education Commission (No.KJZD-K202300603); Project of Chongqing Technological Innovation and Application Development (No.CSTB2022TIAD-GPX0014); Sichuan Science and Technology Program (No.2024JDHJ0038,2024ZYD0140).

}
\thanks{Corresponding author: Tao Wu.}
}

\markboth{IEEE Transactions on Industrial Informatics}
{Xian \MakeLowercase{\textit{et al.}}: GISExplainer: On Explainability of Graph Neural Networks via Game-theoretic Interaction Subgraphs}
\IEEEtitleabstractindextext{%
\begin{abstract}
\justifying
Explainability is crucial for the application of black-box Graph Neural Networks (GNNs) in critical fields such as healthcare, finance, cybersecurity, and more. Various feature attribution methods, especially the perturbation-based methods, have been proposed to indicate how much each node/edge contributes to the model predictions. However, these methods fail to generate connected explanatory subgraphs that consider the causal interaction between edges within different coalition scales, which will result in unfaithful explanations. In our study, we propose GISExplainer, a novel game-theoretic interaction based explanation method that uncovers what the underlying GNNs have learned for node classification by discovering human-interpretable causal explanatory subgraphs. First, GISExplainer defines a causal attribution mechanism that considers the game-theoretic interaction of multi-granularity coalitions in candidate explanatory subgraph to quantify the causal effect of an edge on the prediction. Second, GISExplainer assumes that the coalitions with negative effects on the predictions are also significant for model interpretation, and the contribution of the computation graph stems from the combined influence of both positive and negative interactions within the coalitions. Then, GISExplainer regards the explanation task as a sequential decision process, in which a salient edges is successively selected and connected to the previously selected subgraph based on its causal effect to form an explanatory subgraph, ultimately striving for better explanations. Additionally, an efficiency optimization scheme is proposed for the causal attribution mechanism through coalition sampling. Extensive experiments demonstrate that GISExplainer achieves better performance than state-of-the-art approaches w.r.t. two quantitative metrics: Fidelity and Sparsity.

\end{abstract}

\begin{IEEEkeywords}
Graph Neural Networks, Explanation Methods, Causal Attribution Mechanism, Explanatory Subgraph, Shapley Value.
\end{IEEEkeywords}}

\maketitle

\IEEEdisplaynontitleabstractindextext

\IEEEpeerreviewmaketitle

\IEEEraisesectionheading{\section{Introduction}\label{sec:introduction}}
\IEEEPARstart{G}{raph} neural networks (GNNs) \cite{wu2020comprehensive} have been widely used in modeling graph structured data and achieved remarkable results in a variety of applications such as healthcare, finance, cybersecurity, and more. However, the black-box nature of GNNs prevents users from understanding and trusting the models, thus raising concerns about their reliability \cite{dai2024comprehensive}. Therefore, the explainability of GNNs that aims to answer questions like ``Why did the GNNs model make this prediction?'' is non-trivial. In this context, the focus of this paper is on post-hoc, outcome-oriented, and model-agnostic explanations, which are aimed at explaining any trained black-box models by examining the relationship between input features and a specific prediction of individual instance at the local level \cite{speith2022review}. To this end, feature attribution techniques are popular choices used to explain the model's predictions. Give an input graph, the attribution methods \cite{sanchez2020evaluating} assign importance scores to each input feature (i.e., node or edge) that reflects the contribution of the feature to model's prediction, and then selects the salient edges as an explanatory subgraph.

To identify the input features that are important for specific predictions in GNNs, various explanation methods from different views have been proposed. Based on how the importance scores of input features are calculated, these methods can be classified into four categories: gradients-based methods \cite{pope2019explainability}, surrogate methods \cite{huang2022graphlime}, decomposition methods \cite{schnake2021higher}, and perturbation-based methods \cite{funke2022zorro}. Nonetheless, these methods are prone to explain GNNs based on important nodes or edges, neglecting subgraph-level explanations, which are the building blocks of real-world graphs and closely related to their functionality. To this end, GNNExplainer \cite{ying2019gnnexplainer} was proposed to identify the explanatory subgraph of GNNs by taking edges that give the highest mutual information with the prediction probability. PGExplainer \cite{luo2020parameterized} constructed random graphs by selecting mutually conditionally independent edges from the original input graph, and viewed the random graph that maximizes the mutual information with GNNs' prediction as the explanatory subgraph. PGM-Explainer \cite{vu2020pgm} generated perturbed graphs by randomly disturbing the features of the neighborhood of target node, and then eliminated unimportant features to train Bayesian network for GNNs explanation. From a cause-effect standpoint, Causal Screening \cite{Causal2021} and Gem \cite{lin2021generative} were proposed to measure the causal contribution of an edge on the target prediction and distill the top-$K$ most relevant edges as the explanatory subgraph. Recently, instead of using mutual information \cite{ying2019gnnexplainer, luo2020parameterized, Causal2021} and loss function \cite{lin2021generative} for measuring the importance of input features in feature attribution, Shapley values have been applied to the explanation of GNNs because of their strong theoretical properties \cite{yuan2021explainability}.

\begin{figure}[!t]
\centering
\includegraphics[width=1\linewidth]{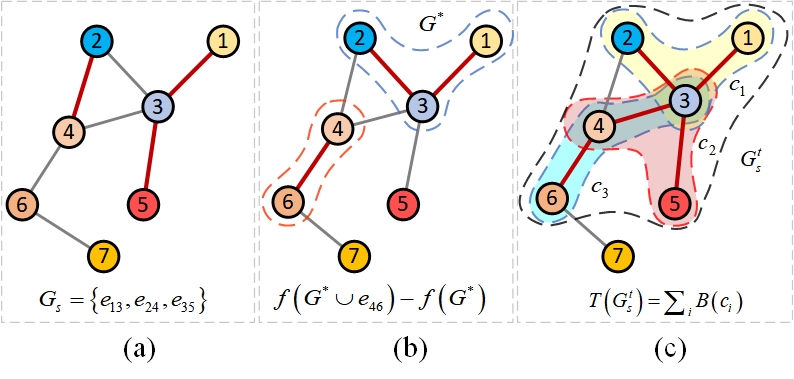}
\caption{Explainability of GNNs. (a) Top-K most relevant edges for explanatory subgraph construction. (b) Causal effect of an edge on model's predictions. (c) The interaction strength of candidate explanatory subgraph based on game-theoretic interaction of multi-granularity coalitions. }
\label{framework}
\end{figure}

By identifying explanatory subgraph for target prediction, the recently proposed explanation methods for GNNs mentioned above have achieved state-of-the-art performance. However, the methods still face the following challenges: (1) \textit{Disconnected explanatory subgraphs are counterintuitive and unfaithful}. Specifically, traditional methods \cite{ying2019gnnexplainer,luo2020parameterized, Causal2021, lin2021generative} select important edges to form an explanatory subgraph for model interpretation, which cannot guarantee the connectivity of the explanatory subgraph. As shown in Fig. 1 (a) and (b), the current salient edge is chosen from the pool of edge candidates, and it may not necessarily be connected to the already identified significant edges. This is inconsistent with the functionality of the graphs, such as molecular structure, and it also contradicts the information aggregation mechanism of GNNs. Furthermore, because there may be negative interactions among the edges, simply converting the edge importance to the subgraph importance for model explanation is unreliable. (2) \textit{Existing explanation methods overlook the interactions between edges, which often form the basis of the aggregation mechanisms in GNNs}. Up to now, the existing explanation methods for GNNs may focus on the importance or causality of edges, as shown in Fig. 1 (a) and (b), but neglect the interactions between them. In fact, different combinations of input edges in graphs, as shown in Fig. 1 (c), often have varying contributions to model predictions \cite{wang2020model, besta2022motif}. (3) \textit{The negative effects of input features on model predictions are ignored by existing methods}. The state-of-the-art explanation methods for GNNs incrementally select nodes or edges that lead to the updated explanatory subgraphs having a greater impact on model predictions. However, this operation may include edges with negative impacts into new explanatory subgraphs, which has not been considered in existing methods.

To address the aforementioned challenges, we introduce the game-theoretic interaction mechanism equipped with causal screening strategy. Although game-theoretic interaction has been considered in recent works of other domains \cite{zhang2021interpreting}, they have not yet been utilized in GNNs and lack the capability to provide explanations for GNNs. Early investigations into real-world graphs have unveiled that recurring subgraphs, i.e., motifs, are the basic modules that control the response to external stimuli, and the importance of nodes or edges can be characterized in terms of their motif participation \cite{Roy2023}. Meanwhile, according to the aggregation mechanism of GNNs, the structural roles of nodes or edges depend on their local connectivity and subgraph patterns \cite{2022Self, 2020GraLSP}. Therefore, the edges in input graphs do not function independently; rather, they interact with other edges to form substructures, known as coalitions of edges, to play their roles. Consequently, analyzing the interactions between edges within different edge coalitions with game-theoretic interaction is crucial for understanding GNNs.

In this paper, we propose a novel explanation method, GISExplainer, that employs game-theoretic interaction for feature attribution in GNNs. GISExplainer models the interactions among multiple input features, i.e, edges, based on game theory, treating each input feature as a player. By defining strongly interacted edges as a coalition, GISExplainer assumes that the edges in the input graph collectively influence the model's prediction within different granularity levels of coalitions, as shown in Figure 1 (c), such that multiple connected salient edges are ultimately recognized as meaningful explanatory subgraph. In detail, GISExplainer consists of three components: edge causal screening mechanism, interaction intensity calculation mechanism, and explanatory subgraph construction mechanism. Specifically, The edge causal screening mechanism selects salient edges and connects them to the previously selected subgraph successively, thereby forming an explanatory subgraph. Interaction intensity calculation mechanism calculates the contribution of candidate explanatory subgraphs to quantify the causal effect of an edge on model's predictions by integrating both positive and negative effects across various coalitions. Explanatory subgraph construction mechanism determines the search strategy for salient edges within the computation graph to ensure the connectivity of the resulted explanatory subgraph. To enhance the computational efficiency of the interaction intensity calculation mechanism, a sampling mechanism for coalitions has been proposed. Overall, the key contributions of this paper are summarized as follows:

\begin{itemize}
  \item A subgraph importance measurement method based on game-theoretic interactions. The method encompasses positive and negative interaction effects of potential multi-grained coalitions in candidate explanatory subgraph to quantify the causal effect of edges. 



  \item A heuristic search strategy for explanatory subgraph construction. This strategy uses the causal contribution of edges as a heuristic function, and combines a causal screening mechanism to iteratively identify candidate edges and select salient edges for expanding the explanatory subgraph.



 \item The efficiency of GISExplainer is optimized based on the connectivity constraints of the explanatory subgraph, and its limitations are analyzed and discussed.

  \item Extensive experiments on syntheic and real-world datasets demonstrate that GISExplainer yields superior performance than state-of-the-art methods and is capable of producing effective explanatory subgraphs both under normal conditions and adversarial attack scenarios.


\end{itemize}

The rest of this paper is organized as follows. Section 2 discusses related works. Section 3 presents the problem definitions and preliminaries. We then elaborate the proposed framework in Section 4. Experimental setup and discussion of results are provided in Section 5. Finally, we conclude the paper and give future directions.

%
%

\section{Related Work}
In this section, we review the works strongly related to our work, including explainability in deep neural networks, explainability in GNNs and game theory based model explainability.




\subsection{\label{sec:level1} Explainability in Deep Neural Networks }
Considerable research efforts have been devoted to the post-hoc explanation of deep neural networks (DNNs). From a holistic perspective, model distillation achieves explanation of DNNs by training a smaller model, often referred to as a student model, to mimic the decision-making behavior of a complex target model, known as a teacher model \cite{frosst2017distilling}. Furthermore, activation maximization adjusts input data to optimize the model's prediction probabilities, thereby disclosing what a neural network perceives as important features for a specific class \cite{yoshimura2021toward}. From a local perspective, there are mainly four types of explanation methods to answer why a DNN model makes a specific prediction on a particular test sample: sensitivity analysis, local approximation, back propagation, and class activation mapping. Specifically, sensitivity analysis utilizes the gradient of the model's prediction with respect to the input features as an explanation, such as Saliency Map \cite{petsiuk2021black} and SmoothGrad \cite{smilkov2017smoothgrad}. Local approximation methods utilize explainable models with simple architecture to fit the prediction process of the target model for specific instance. For example, LIME \cite{ribeiro2016should} explains the predictions of any black box classifier for individual instances by approximating it locally with an explainable model. Backpropagation methods propagate the model's decision signals from the output layer to the input layer of the model to deduce the significance of input features. Examples include Grad-CAM \cite{selvaraju2017grad}, LRP\cite{bach2015pixel}, DeepLIFT\cite{li2021deep}, and others. Class activation mapping methods like CAM \cite{zhou2016learning} and Grad-CAM \cite{selvaraju2017grad} determine the discriminative regions of an image that are important to the prediction of a particular class. It's worth noting that the above methods are primarily based on statistical interpretations and assume that input features are independent of each other, making it difficult to capture the interactions between edges in GNNs.

\subsection{\label{sec:level1} Explainability in Graph Neural Networks}
With the tremendous success of GNNs in modeling graph data, the explainability of GNNs has attracted increasing attention. To identify the input features that significantly influence the predictions of GNNs, the most straightforward solution is to utilize the gradients of target prediction with respect to input features, i.e., gradients-based methods. Traditional DNNs explanation methods sensitivity analysis and back propagation were applied to GNNs \cite{baldassarre2019explainability}, where higher gradient values indicate more important input features. The explainability method Grad-CAM originally designed for DNNs was extended to GNNs \cite{pope2019explainability}, which utilizes the gradients of the target prediction with respect to the features as class-specific weights and then generates heat-maps through weighted summation based on these weights. Next, from the perspective of the impact of input perturbations on model predictions, besides the methods GNNExplainer \cite{ying2019gnnexplainer}, PGExplainer \cite{luo2020parameterized}, PGM-Explainer \cite{vu2020pgm}, and Causal Screening \cite{Causal2021} introduced in the Introduction, GraphMask \cite{schlichtkrull2020interpreting} was proposed for interpreting the predictions of GNNs by identifying unnecessary edges. RCExplainer \cite{wang2022reinforced} jointly considers causal screening strategy and reinforcement learning agent to pick up salient edge at each step for explanatory subgraph construction. SubgraphX \cite{yuan2021explainability} explores different subgraphs with Monte Carlo tree search and identified important subgraphs using Shapley values to explain GNN predictions. Perturbation-based methods also include TAGExplainer \cite{xie2022task}, CF$^2$ \cite{tan2022learning}, CLEAR \cite{ma2022clear}, and others. Moreover, in order to approximate the complex GNNs with a simple and interpretable surrogate model, GraphLIME \cite{huang2022graphlime} learns a nonlinear explanation model locally in the subgraph of the node being explained. ILS \cite{heidari2023explaining} utilizes group sparse linear models as local surrogates to approximate the behavior of a black-box GNNs. DnX \cite{pereira2023distill} uses knowledge distillation to learn a global surrogate model to mimic the behavior of GNNs. Beyond
that, following decomposition rules, the decomposition methods WB-LRP \cite{li2025wb}, DEGREE \cite{feng2022degree} and GNN-LRP \cite{schnake2021higher} were proposed to distribute the prediction score layer by layer until the input layer.

However, extensive studies have demonstrated that real-world graphs exhibit the recurrence of subgraphs, called motifs \cite{milo2002network, benson2016higher}. Edges within motifs often work together with other edges, forming a coalition of edges, to influence the model's predictions. Despite the success of explainability in GNNs, these methods fail to consider the coalition of edges.

\subsection{\label{sec:level1} Game Theory based Model Explainability}

Game theory provides a powerful framework for analysing the strategic interactions of individuals in strategic situations, where the ultimate results depend on the choices of all the players involved. Recently, the game theoretic concept Shapley value is increasingly being applied to the explainability of DNNs. Specifically, in order to develop attribution methods with strong theoretical guarantees, a perturbation based method deep approximate shapley propagation (DASP) \cite{ancona2019explaining} was proposed for DNNs explanation. Based on game-theoretic Shapley value, TreeExplainer \cite{lundberg2020local} was proposed as an explanation method for trees that enables the tractable computation of optimal local explanations. By taking into account the causal relationships of input features, causal Shapley value \cite{heskes2020causal} was designed based on causal chain graphs. Moreover, based on Shapley value, SubgraphX \cite{yuan2021explainability}, MAGE \cite{bui2024explaining} and GT-CAM \cite{li2024gt} were proposed for the explainability of GNNs. However, the above methods assume that each player participates in the game and receives a reward individually, while the interaction among multiple players is not taken into account. In recent years, the concept of game-theoretic interaction has been proposed to measure the overall utility of interactions between players \cite{zhang2021interpreting, zhou2024interpretability}. However, it cannot be directly applied to the explanation of GNNs.



\section{Problem Definition and Preliminaries}
\subsection{\label{sec:level1}Problem Definition}

\begin{definition} [Input Graph] Let ${\mathcal G} = \left( {{\mathcal{V}}{\rm{,}}{\mathcal E}} \right)$ represents a undirected and unweighted input graph, where ${\mathcal{V}} = \left\{ {{v_1},...,{v_N}} \right\}$ denotes the set of nodes with size $N$, ${\mathcal E} = \left\{ {{e_1},...,{e_M}} \right\}$ denotes the set of edges with size $M$. Formally, we denote the adjacency matrix of graph ${\mathcal G}$ as $\mathbf{A} \in {\mathbb{R}^{N \times N}}$, in which ${{\mathbf{A}}_{ij}} = 1$ if ${v_i}$ and ${v_j}$ are connected in ${\mathcal G}$, otherwise ${{\mathbf{A}}_{ij}} = 0$. Each edge ${e_i}$ can be represented as a node pair $(u, v)$, where $u, v \in \mathcal{V}$. Let ${\cal N}(u)$ denote the neighbors of node $u$, ${\cal N}(u) = \{ v|(u,v) \in {\cal E}\}$.
\end{definition}

\begin{definition} [Computation Graph] The computational graph $ {\mathcal G}_c(v) $ for a node  $ v $ encompasses all the data required by the GNN model $ f( \cdot ) $ to make a prediction $ \hat{y} $ at that node, which is determined by the aggregation mechanism of GNNs. In the case of a two-layer GNN model, this graph comprises neighbors within two hops.
\end{definition}

\begin{definition} [Explanatory Subgraph] Generally, the messages in GNNs are transmitted and aggregated between nodes in a graph along the edges that connect them, and the computational graph $ {\mathcal G}_c(v) $ includes edges that are crucial for prediction $ \hat{y} $ as well as those that are less important. Consistent with previous studies in the literature \cite{lin2021generative, wang2022reinforced}, our attention is centered on providing explanations on graph structures. Thus, the explanatory subgraph only contains the most relevant edges $e_i^*$ of the computational graph $ {\mathcal G}_c(v) $ for the prediction, i.e., ${G_s}(v) = \{ e_1^*,e_2^*,...,e_K^*\}  \subseteq {G_c}(v)$.
\end{definition}

\begin{definition} [GNNs Explainability] Given a GNN model $f(\cdot)$ for node classification, explainability of GNNs aims to obtain an explanation model, denoted as ${f_{\exp }}( \cdot )$, that can decompose the prediction $ \hat{y} $ into input features and calculate an importance score $S( \cdot )$ for each feature, indicating how much it contributes to the prediction. Based on this, an explanatory subgraph ${G_s}(v)$ can be obtained, which can provide faithful explanation for the prediction $ \hat{y} $. In particular, we do not require any prior knowledge about the model's architecture and parameters. We allow the explainers to obtain prediction results by performing queries the target model $f(\cdot)$. We require that the obtained explanatory subgraph ${G_s}(v)$ is connected.
\end{definition}

\begin{figure*}[!htb]\centering
	\centering
	{ \includegraphics[width = 7.5in]{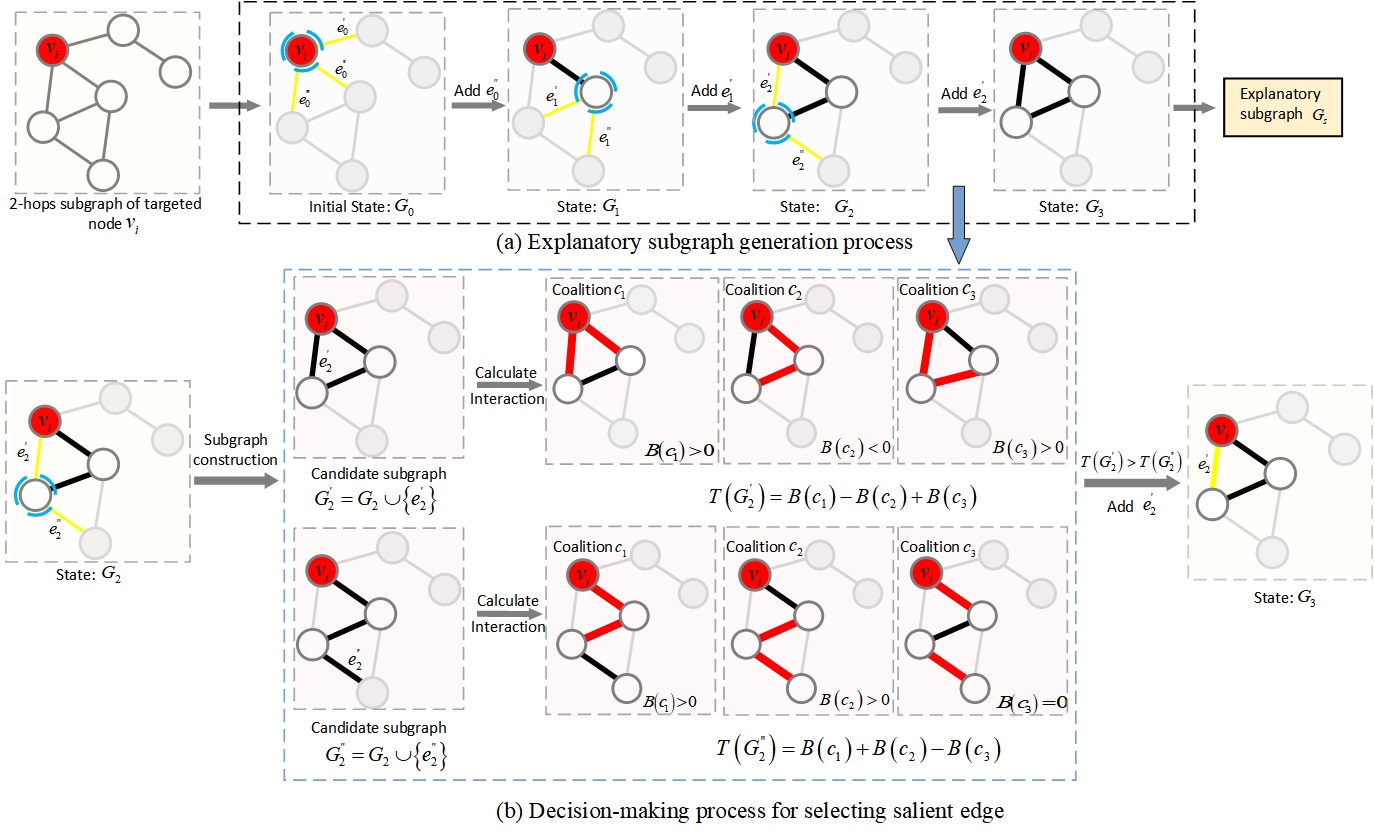}}
	\caption{Illustration of the proposed GISExplainer. (a) The process of generating an explanatory subgraph progressively. Here, the red node represents the target node $v_i$, and the nodes with blue dashed coil is the extension node. The yellow edges indicate the candidate edges that can be selected for updating the explanatory subgraph. (b) The decision-making process of causal screening for salient edges. The salient edge is selected for updating the explanatory subgraph based on the contributions of the candidate edges, which can be measured by the interaction strength of the corresponding explanatory subgraphs. $B(\cdot)$ indicates that the reward of the coalition, and the interaction strength $T(G_i)$ of a corresponding explanatory subgraph is the combined effects of the coalitions. }
	\label{framework}
\end{figure*}


\subsection{Graph Neural Networks}
GNNs are a powerful class of neural networks that can effectively model and analyze complex relationships and dependencies within graph data. They utilize a message-passing scheme to propagate information between neighboring nodes in the graph, enabling them to capture both local and global patterns. The core computation of GNNs can be described as follows:

\begin{equation}
	\label{gnn}
	\bf{H}_v^{(l+1)} = \sigma \left(\sum_{u \in \mathcal{V}(v)} \frac{1}{ c_{u,v} } \bf{W}^{(l)} \bf{H}_u^{(l)} \right),
\end{equation}
where $\bf{H}_v^{(l+1)}$ represents the hidden representation of node $v$ at layer $l+1$, $\mathcal{V}(v)$ denotes the set of neighbors of node $v$, $\bf{W}^{(l)}$ is the weight matrix at layer $l$, $\sigma(\cdot)$ is the activation function, and $c_{u,v}$ is a normalization factor. Equation (1) captures the iterative process of message passing and aggregation, allowing GNNs to capture and integrate information from the neighborhood of each node, producing informative and context-aware node representations.

\subsection{Shapley Value}
The Shapley value is a solution concept in cooperative game theory \cite{shapley1953value}. It describes how to fairly distribute the total gains of a game to the players depending on their respective contribution, assuming that they all collaborate \cite{duval2021graphsvx}. Given a game $g( \cdot )$ with $N$ players $P = \{ {p_1},{p_2},...,{p_n}\}$, where each player $p_i$ can participate in the game $g( \cdot )$ individually or form a coalition $S$, $S \subseteq P$, with other players to participate in the game as a group, the Shapley value is based on the marginal contribution of each player to all possible coalitions of players. The function $g( \cdot )$ is a characteristic function that maps subsets of players to the real numbers ${2^N} \to \mathbb{R}$, with $g(\emptyset ) = 0$, where $\emptyset$ denotes the empty set. Consequently, $g(S)$ describes the total expected sum of payoffs the members of coalition $S$ can obtain by cooperation. According to the Shapley value, the contribution of player $i$ can be formalized as follows:

\begin{equation}
	\varphi_g(p_i|P) = \sum_{S \subseteq P \setminus{p_i}} \frac{(n - |S| - 1)!|S|!}{n!} [g(S \cup {p_i}) - g(S)],
\end{equation}
where $P$ represents the set of all players, $\varphi_g(p_i|P)$ is the Shapley value of player $p_i$, $|S|$ denotes the number of players in $S$, $n$ represents the total number of players. The Shapley value ranges between \([-1, 1]\). If the value is positive, it indicates that the current edge has a positive impact on the model's prediction; conversely, a negative value suggests a negative impact. The magnitude of the it represents the strength of player's influence. In this paper, we use $\varphi(p_i|P)$ to represent $\varphi_g(p_i|P)$ in the following manuscript for simplicity.

\subsection{Game-theoretic Interaction}
It's worth noting that the Shapley value merely measures the average added worth, i.e., the average of the difference of worth $g(S \cup {p_i}) - g(S)$, that player $p_i$ brings to all possible coalitions, and it gives no information on the phenomena of interaction or cooperation existing among players  \cite{shapley1953value}. Assuming players $p_i$ and $p_j$ often participate in or abstain from a game together, thereby forming a coalition. This coalition can be considered as one singleton player, denoted by ${S_{ij}}$, and the players of the game becomes $P' = P\backslash \{ {p_i},{p_j}\}  \cup {S_{ij}}$. The reward obtained by this coalition is always different from the sum of the rewards that players $p_i$ and $p_j$ would obtain if they participated in the game individually. Thus, according to the Shapley value defined in equation (2), the interaction between players $p_i$ and $p_j$ can be defined as the additional reward generated by the coalition with respect to the sum of the rewards of each player participating the game individually \cite{zhang2021interpreting}, i.e.,

\begin{equation}
B({S_{ij}}) = \varphi ({S_{ij}}|P') - [\varphi ({p_i}|{P_i}) + \varphi ({p_i}|{P_i})],
\end{equation}
where the set of players ${P_i} = P\backslash \{ j\}$, ${P_j} = P\backslash \{ i\}$.

We can extend the definition of interaction to multiple players, for any set of players \( S \subseteq P \),  in which the players always participate in the game together and form a coalition. Thus, they can be treated as a singleton player, denoted by $[S]$. The game-theoretic interaction of the coalition $B([S])$ can be formulated as:

\begin{equation}
	\label{interaction index}
	B([S]) = \varphi([S]|P[S])-\sum_{p_i\in S}\varphi(p_i|P_i),
\end{equation}
where $\varphi ([S]|P[S])$ represents the Shapley value of the coalition $[S]$, the player set $P[S] = P\backslash S \cup \{ [S]\}$, and $\varphi(p_i|P_i)$ denotes the Shapley value of player $p_i$ in coalition $S$.

%

\section{The Proposed Method}

This section presents the proposed GNNs explanation method GISExplainer. Fig. 2 illustrates the process of progressively generating the explanation subgraph and the decision-making process for selecting salient edges. Fig. 2 (a) illustrates the explanation subgraph construction process based on causal screening. GISExplainer starts with an empty set centered on target node $v_i$ and updates the explanatory subgraph at each step by selecting the edge that can maximize the game-theoretic interaction of the candidate explanatory subgraphs. The specific causal screening strategy will be detailed in Section 4.1. Fig.2 (b) provides a detailed illustration of the decision-making process of causal screening for salient edges at state 2. For the current explanatory subgraph, multiple candidate edges are added to form corresponding new candidate explanatory subgraphs. The interaction strength of each candidate explanation subgraph is then calculated to measure the causal contribution of the edges. Thus, the edge with the maximum causal contribution among them is considered the salient edge at the current stage, and its corresponding candidate explanatory subgraph becomes the new explanatory subgraph of state 3. The specific steps of the decision-making process will be described in Section 4.2. The above process is repeated according to the graph search strategy until a termination condition is met, which will be illustrated in in Section 4.3.

\subsection{Causal Screening for Salient Edges}

The generation process of an explanatory subgraph can be seen as a sequential decision process that considers the causal contribution of edges and incrementally updating explanatory subgraph by selecting a salient edge at each step. To the computation graph ${\mathcal G}_c$ for node ${v_i}$, the explanatory subgraph at state $i$ is $G_s^i$, and the candidate edge set is $ \mathcal{O}_k $ obtained by the graph search strategy described in section 4.3. Therefore, the question needed to be answered is, "Which candidate edge in the set $ \mathcal{O}_k $ should be used as salient edge to update the explanatory subgraph in the current state?". According to causal analysis \cite{Causal2021}, we can obtain the salient edge by maximizing the causal contribution of edges, which can be formulated as follows:

\begin{equation}
    e_k^* = \arg \mathop {\max }\limits_{{e_k} \in {\mathcal{O}_i}} C({e_k}|G_s^i),\;k = 1,2,...,K,
\end{equation}
where $e_k^*$ is the salient edge selected from the candidate edge set $ \mathcal{O}_i $ at step $i$. By integrating the explanatory subgraph $G_s^i$ with $e_k^*$, a new explanatory subgraph $G_s^{i+1}$ can be obtained. Therefore, by repeating the above process until no candidate edges is available for updating, and the final explanatory subgraph for the model prediction can be obtained. To instantiate the causal contribution of edge, we compare the importance of different subgraphs. Specifically, the calculation of the causal contribution of edges can be defined as follows:
\begin{equation}
C({e_k}|G_s^i) = T(G = G_s^i \cup \{ {e_k}\} ) - T(G = G_s^i),
\end{equation}
where $T(G)$ represents the importance of subgraph $G$.

The basic idea of GISExplainer is integrating the screening strategy with the importance of subgraphs. Here, the explanation subgraph starts from an empty set and successively selects edges that maximize the importance scores of the explanatory subgraph $G_s^i$ merging the candidate edge $e_k$. Formally, the objective function is formulated as follows:
\begin{equation}
e_k^* = \arg \mathop {\max }\limits_{{e_k} \in {O_i}} (T([G_s^i \cup \{ {e_k}\} ]) - T([G_s^i])),\;k = 1,2,...,K,
\end{equation}
where $e_{k}^*$ is the edge that maximizes the importance of the newly formed subgraph. The candidate edge set $ \mathcal{O}_i $ consists of edges that are connected to the edges of the previous explanatory subgraph $G_s^i$. $G_s^i \cup \{ {e_k}\}$ represents the newly generate candidate explanatory subgraph formed by adding the candidate edge $ e_k $ to the previous explanatory subgraph $G_s^i$. $T([G_s^i \cup \{ {e_k}\} ]$ represents the importance of the new subgraph. By calculating the difference between the importance of the new subgraph $G_s^i \cup \{ {e_k}\}$ and that of the previous one $G_s^i$, the edge that results in the greatest increase in the importance for the new subgraph is selected.

\subsection{Game-theoretic Interaction for Subgraph Importance}

To calculate the importance $T([G_s^i])$ of subgraph $G_s^i$, we reformulate the concepts of Shapley value and game-theoretic interaction of the coalition defined in equation (2) and (4) within the context of graph model explanation. Specifically, we regard the prediction of target GNN model as a game, and treat the edges in a subgraph relevant to target node $v_i$ as players. Let $E$ denote the set of edges included in the subgraph, $n$ represents the number of edges, $E'$ is a subset of $E$, and $|E'|$ denotes the size of $E'$. Thus, the Shapley value can be rewritten as follows:

\begin{equation}
        \label{shaplye value}
	\varphi(e_k|E) = \sum_{E' \subseteq E\setminus{e_k}} \frac{(n - |E'| - 1)!|E'|!}{n!} [g(E' \cup {e_k}) - g(E')].
\end{equation}
For target model with multiple node category, $g(\cdot)$ is implemented as the confidence score of the true category.

To measure the collective contribution of multiple edges in GNNs, we employ the game-theoretic interaction of multiple players. Suppose the set of edges $ A \subseteq E $ forms a coalition; we regard this coalition as a singleton player \([A]\) participating in the game. According to the equation (4), the game-theoretic interaction of coalition \([A]\) can be defined as follows:
\begin{equation}
	\label{gnninteraction}
	B([A]) = \varphi([A]|E[A])-\sum_{e_i\in A}\varphi(e_i|E_{i}),
\end{equation}
where $\varphi([A]|E[A])$ represents the Shapley value of the coalition $[A]$. $E[A]$ is the player set of the game, $E[A]= E\setminus A \cup \{[A]\}$. Similarly, $\varphi(e_i|E_{i})$ denotes the Shapley value of edge $e_i$ when the model prediction is produced over the player set $E_{i}$, $E_{i} = E\setminus{e_i}$.

\begin{figure}[!htp]
\centering
\includegraphics[width=0.88\linewidth]{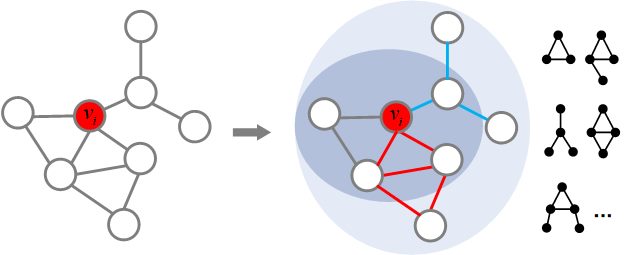}
\caption{Coalitions for game-theoretic interaction of subgraphs. }
\end{figure}

Early studies in the field of complex networks suggested that real-world graphs are not homogeneous, and substructures, or motifs, are the basic building blocks and functional unit of them \cite{milo2002network, sporns2004motifs}. In particular, substructures can combine with each other to form hierarchical organization of real-world graphs, and the substructures that follows a specific structural pattern can be represented by each other for network reconstruction. Moreover, the substructures always correspond to the functional group in networks such as molecular graphs and brain networks, enabling them to respond to external stimuli. Along this line, the model weights learned in GNNs correspond to the edges in the graphs and reflect the structural patterns of them. That is, the computation graph of GNNs includes multiple substructures with varying contributions, as shown in Fig. 3. Therefore, the explainability of GNNs, which identifies explanatory subgraphs from the computation graph, requires considering the significance of all potential substructures, i.e., the coalitions of edges in game scenario. Here, given a subgraph $G_s^i$ and the coalitions $A_i$ in it, ${A_i} \subseteq G_s^i$, the importance of the subgraph can be defined based on the absolute values of the game-theoretic interaction of all coalitions within it:

\begin{equation}
T([G_s^i]) = \sum\nolimits_{{A_i} \subseteq G_s^i,|{A_i}| > 1} {|B({A_i})|}.
\end{equation}

To explain the prediction results of GNNs regarding specific instances, it is crucial not only to consider increasing model confidence and identifying coalitions with positive effect, but also to recognize the importance of finding coalitions that reduce model confidence and exert negative effect. Therefore, the importance of subgraphs needs to take into account both the positive and negative effects of coalitions. Here, we define the positive effects of all coalitions in subgraph $G_s^i$ as follows:
\begin{equation}
{T^ + }([G_s^i]) = \sum\nolimits_{{A_i} \subseteq G_s^i, B({A_i}) >0, |{A_i}| > 1} {B({A_i})}.
\end{equation}
Similarly, the negative effects of all coalitions in the subgraph are defined as ${T^ - }([G_s^i]) = \sum\nolimits_{{A_i} \subseteq G_s^i, B({A_i}) <0, |{A_i}| > 1} {B({A_i})}$. Therefore, the importance of the subgraph defined in equation (10) can be rewritten as follows:

\begin{equation}
T([G_s^i]) = {T^ + }([G_s^i]) - {T^ - }([G_s^i]).
\end{equation}
It is evident that $T([G_s^i])$ is the result of subtracting a negative value from a positive value, leading to a positive outcome.

\renewcommand{\algorithmicrequire}{\textbf{Input:}}
\renewcommand{\algorithmicensure}{\textbf{Output:}}
\begin{algorithm}[!htb]
\caption{Explanatory subgraph-oriented heuristic search}
\label{alg::conjugateGradient}
\begin{algorithmic}[1]
\REQUIRE Computation graph $ {\mathcal G}_c(v) $, target node $v$.
\ENSURE Explanatory subgraph ${\mathcal G}_s^*$.

\STATE {\bfseries Initialization:} Use target node $v$ as extension node, adopt the causal contribution of edges defined in equation (7) as a heuristic function, define an empty edge queue $Q = \emptyset$, $i = 0$.
\STATE {\bfseries Expand Edges:} Define all neighboring edges of the extension node as candidate edges $ \mathcal{O}_i $. Evaluate all the candidate edges and assign each edge a value based on the heuristic function.
\STATE {\bfseries Select Salient Edge:} From the candidate edges $ \mathcal{O}_i $, select the edge $e_k^*$ with the highest heuristic value.
\STATE {\bfseries State Update:} Update edge queue $Q = Q \cup e_k^*$, and use the new end node of edge $e_k^*$ as the extension node. $i = i+1$, and update the candidate edges $ \mathcal{O}_i $ using all the neighboring edges of it.
\STATE {\bfseries Goal Check:} If the candidate edge set $ \mathcal{O}_i $ is empty, or $T([Q]) < T([Q\backslash e_k^*])$, terminate the search.
\STATE {\bfseries Continue:} If not, repeat steps 2 to 5 for the next salient edge until the goal is reached.
\STATE {\bfseries Return:} The subgraph constructed based on the edge queue $Q$ is returned as explanatory subgraph ${\mathcal G}_s^*$.

\end{algorithmic}
\end{algorithm}

\subsection{Heuristic Search for Explanatory Subgraph}
To construct an explanatory subgraph for target GNN model $f( \cdot )$, at each step of the sequential decision process, the edge with the highest causal contribution is selected from the candidate edges for updating the previous explanatory subgraph. Inspired by greedy best-first search \cite{ferber2022learning}, this paper models the aforementioned problem as a graph search problem and proposes a heuristic search strategy guided by the causal contribution. Specifically, on the basis of computational graph $ {\mathcal G}_c(v) $, the search strategy uses the causal contribution of edges defined in Equation (7) as a heuristic function to determine which edge is the most promising, and evaluates the cost of each possible edge and then expanding the edge with the highest heuristic value. The process is repeated until the candidate edge set $ \mathcal{O}_i $ is empty or the subgraph's importance score $T([E])$ is no longer increasing. The complete process of the heuristic search strategy for explanatory subgraph is shown in Algorithm 1.

Fig. 4 provides an example to illustrate the heuristic search strategy for explanatory subgraph construction. Here, the node with red color is the target node. At each step, the node with a blue dashed coil is the extension node, the edges with yellow color are the candidate edges $ \mathcal{O}_i $, and the black bold edges form the current explanatory subgraph. The process of constructing the explanatory subgraph involves sequentially selecting salient edges from a set of candidate edges and updating the previous explanatory subgraph accordingly.

\begin{figure}[!t]
\centering
\includegraphics[width=0.9\linewidth]{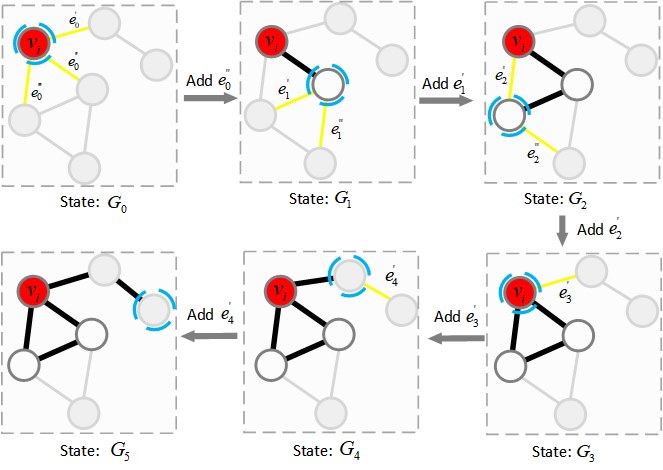}
\caption{Heuristic search for explanatory subgraph construction. }
\end{figure}

\section{Analysis and Discussion}

\subsection{Efficiency Enhancement and Limitations}
The importance of the explanatory subgraph defined in Equation (10) requires the calculation of the game-theoretic interaction of all potential coalitions in it. On the one hand, given the $L$-layer target GNN model, the feature representation of the target node is refined by aggregating information from its $L$-hop computational graph. Thus, the number of edges in explanatory subgraph and the coalitions they form is limited, which alleviate computational load of our method GISExplainer. On the other hand, we recognize that calculating Shapley values and game-theoretic interaction requires considering the marginal contributions over all possible combinations of edges, the number of which directly impacts computational efficiency. Meanwhile, disconnected explanatory subgraphs are counterintuitive and may lead to unfaithful explanations, thus this paper requires that the explanatory subgraph be connected. Thus, we propose to add a connectivity check mechanism to Shapley values and game-theoretic interaction, avoiding unnecessary calculations. Formally, the optimized Shapley value can be defined as follows:

\begin{equation}
\varphi ({e_k}|E) = \left\{ \begin{array}{l}
\sum\limits_{E' \subseteq E\backslash {e_k}} {\zeta (n,E')[\tau ({e_k},E')],E'{\kern 1pt} is{\kern 1pt} connected} \\
0,otherwise
\end{array} \right.
\end{equation}
where $\zeta (n,E') = \frac{{(n - |E'| - 1)!|E'|!}}{{n}}$, $\tau ({e_k},E') = g(E' \cup {e_k}) - g(E')$. Similarly, the optimized game-theoretic interaction can be defined as follows:
\begin{equation}
B([A]) = \left\{ \begin{array}{l}
\varphi ([A]|E[A]) - \sum\limits_{{e_i} \in A} {\varphi ({e_i}|{E_i})} ,A{\kern 1pt} is{\kern 1pt} connected\\
0,otherwise
\end{array} \right.
\end{equation}

It should be noted that, although the heuristic search strategy proposed in this paper is simple, fast, and easy to implement, it has some disadvantages, such as inaccurate results, local optima.

\subsection{Comparison with Related Works}
To provide explanations for GNNs, GNNExplainer \cite{ying2019gnnexplainer} and PGExplainer \cite{luo2020parameterized} directly perturbs the input graph and selecting nodes and edges with the maximum values as output. However, they fail to consider the dependencies between input features and belong to additive feature attribution methods. In contrast, PGM-Explainer \cite{vu2020pgm} characterizes the dependencies of input features based on conditional probabilities, and Gem \cite{lin2021generative}, Causal Screening \cite{Causal2021}, and RC-Explainer \cite{wang2022reinforced} quantify the relationship among input features from a causal perspective. Moreover, from the perspective of importance measuring, different from mutual information based GNNExplainer \cite{ying2019gnnexplainer}, PGExplainer \cite{luo2020parameterized} and Causal Screening \cite{Causal2021}, the decrease in model error based Gem \cite{lin2021generative}, reinforcement learning based RC-Explainer \cite{wang2022reinforced}, and Shapley value bassed SubgraphX \cite{yuan2021explainability}, our method GISExplainer adopts game-theoretic interaction for importance measuring. Additionally, different from the existing methods, our method GISExplainer is the first to consider the contributions of different coalitions in the process of evaluating explanatory subgraphs, and the calculation of interaction intensity takes into account both positive and negative effects.




\section{Experiments}

\subsection{Experimental Setup}

\subsubsection{Target Models and Datasets}

Here, to validate the effectiveness of our model, a two-layer Graph Convolutional Network (GCN) \cite{gcn2016semi} and a two-layer Graph Isomorphism Network (GIN) \cite{gIN2018powerful} are used as target models. Our experiments are conducted on a single NVIDIA 2080Ti GPU, powered by an Intel CPU, with the software stack comprising Python 3.7.0, PyTorch 1.1.13, and PyTorch Geometric 2.3.1. Moreover, according to \cite{ying2019gnnexplainer}, four synthetic datasets BA-Shapes, BA-Community, Tree Cycle, and Tree Grids, and three real-world datasets Cora \cite{bandyopadhyay2005link}, Citeseer \cite{sen2008collective}, and Pubmed \cite{mccallum2000automating}, are adopted for performance evaluation. In particular, BA-Shapes dataset is base on Barabasi-Albert (BA) graph and a set of five-node ''house``-structured network motifs, and is further perturbed by adding 10 percent random edges. BA-Community dataset is a union of two BA-SHAPES graphs, and node features and class labels are generated based on their structural roles and community memberships. Tree Cycle dataset is generated based on a 8-level balanced binary tree and 80 six-node cycle motifs. Tree Grids dataset is similar to Tree Cycle, except that they use 3-by-3 grid motifs instead of 80 six-node cycle motifs. The statistical properties of them are shown in Table 1. 

\begin{table*}[!htp]
	\setlength{\tabcolsep}{1.3mm}
	\centering
	\caption{Statistical properties of the datasets.}
    \begin{tabular}{c|cccccccc}
      \hline
      \multirow{1}{*}{Dataset} & \multirow{1}{*}{BA-shapes} & \multirow{1}{*}{BA-community} & \multirow{1}{*}{Tree-cycle}  & \multirow{1}{*}{Tree-grid} & \multirow{1}{*}{Cora} & \multirow{1}{*}{CiteSeer} & \multirow{1}{*}{PubMed}  \\ \hline
      \multirow{1}{*}{Node}   & 700  & 1400 & 871  & 1231 & 2708 & 3327 & 19717 \\  \hline
      \multirow{1}{*}{Edges} & 4110 & 8920 & 1950 & 3410 & 5489 & 4732 & 44338 \\  \hline
      \multirow{1}{*}{Classes} & 4    & 8    & 2    & 2  & 6 & 7 & 3 \\    \hline
	\end{tabular}%
	\label{tab:time-result}%
\end{table*}%

\begin{figure*}[!tbp]\centering
	\centering
	\label{visualization}
	{ \includegraphics[width = 7.5in]{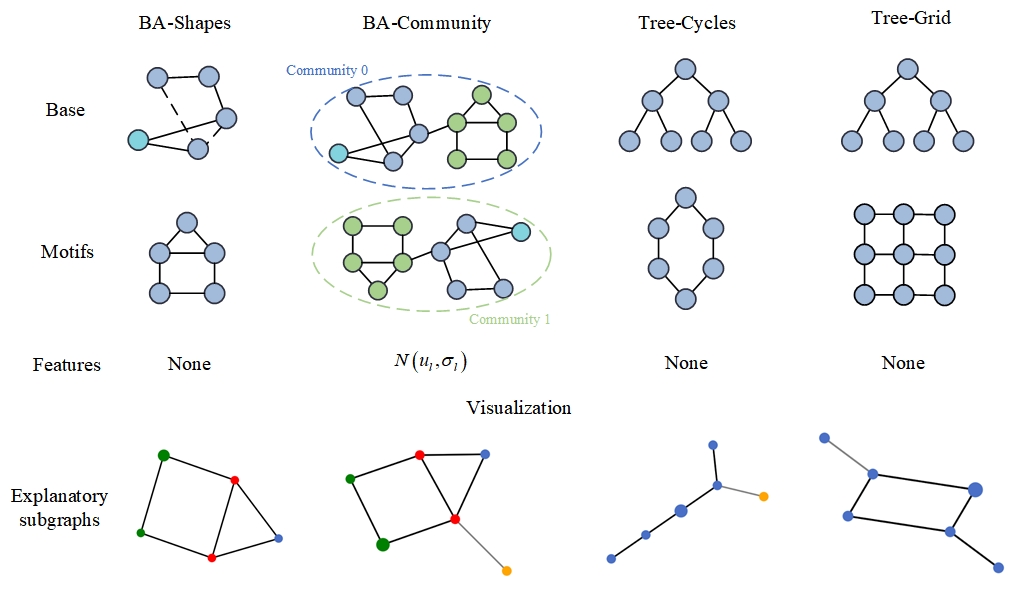}}
	\caption{Visualization of GISExplainer's explanatory subgraphs on synthetic datasets	}
	\label{framework}
\end{figure*}

\subsubsection{Evaluation Metrics}
Fidelity and Sparsity are adopted in this paper, as introduced in \cite{ying2019gnnexplainer}, to evaluate the performance of explanation methods. Given a input graph $\mathcal G$, a target node $v$ and its predicted class $y$, and the explanatory subgraph ${\mathcal G}_s$ that is relevant to node $v$, Fidelity measures the impact of removing significant graph structures connected to node $v$ on the model's prediction for that node. It quantifies the change in the confidence of the predicted class when part of the graph related to node $v$ is excluded. The Fidelity metric can be formulated as follows:

\begin{equation}
	Fidelity = \frac{1}{N}\sum_{i=1}^{N} (f({\mathcal G})_{y_i} - f({\mathcal G} \setminus {\mathcal G}_{S_i})_{y_i})
\end{equation}
where $N$ is the total number of test nodes, and $f(\mathcal G)_{y_i}$ represents the model's predicted confidence for class $ y_i$ on the original graph $ G $ for node $ i$. A high Fidelity score indicates that the removal of the important subgraph $ {\mathcal G}_{S_i} $ connected to node $ i $ leads to a significant change in the model's prediction confidence for that node, suggesting that the explanatory subgraph is crucial for the classification of node $ i $. Additionally, we employ the Sparsity metric to assess the efficiency of the explanations by measuring the proportion of edges selected within the subgraphs associated with each node. A sparser subgraph indicates a more concise and focused explanation. The Sparsity metric is defined as follows:

\begin{equation}
\textit{Sparsity} = \frac{1}{N}\sum_{i=1}^{N} \left(1 - \frac{|{\mathcal G}_{s_i}|}{|G_i|}\right)
\end{equation}
Here, $N $ is the total number of test samples (nodes), $ |{\mathcal G}_{s_i}| $ is the number of edges in the explanatory subgraph $ {\mathcal G}_{s_i} $ specific to node $ i $, and $ |{\mathcal G}_i| $ is the total number of edges in the computation graph relevant to node $ i $. A higher Sparsity value suggests that fewer edges are selected for the explanation, which is preferable for clarity and efficiency.


\subsubsection{Baselines}

We compare GISExplainer with the state-of-the-art explainers, including mutual information based GNNExplainer \cite{ying2019gnnexplainer} and PGExplainer \cite{luo2020parameterized}, conditional probability based PGM-Explainer \cite{vu2020pgm}, model error based Gem \cite{lin2021generative}, Shapley value based SubgraphX \cite{yuan2021explainability}.

\subsection{Experimental Results}
\subsubsection{Experiments on Syntheic Datasets}

A qualitative analysis and evaluation of GISExplainer on synthetic datasets is conducted firstly. Fig. 5 presents the specific explanatory subgraphs obtained on various datasets. Among these identified explanatory subgraphs, the largest node represents the target node being classified. The results indicate that GISExplainer successfully identifies the ''house`` structure within the BA-Shapes and BA-Community datasets. It is important to note that the experiment is tailored to explain the 2-hop GNNs, thus limiting the search for the explanatory subgraph to the 2-hop neighborhood of the target node. This constraint has led to an incomplete recognition of ''cyclic`` and ''grid`` structures in the Tree-Grid and Tree-Cycles datasets by GISExplainer. Despite this, the visualizations demonstrate that GISExplainer has made significant efforts to discern the underlying motif structures. Therefore, the above visualization results can intuitively demonstrate the effectiveness of the explanation method GISExplainer proposed in this paper.

\begin{table*}[!tp]
    \centering
    \label{gcn_syn}
    \caption{Fidelity and sparsity scores for the top 20 nodes in the test set of synthetic datasets for 2-hop GCN explanations. Higher values signify better performance, with bold black denoting the optimal outcome.}
    \label{tab:GCN_scores}
    \setlength{\tabcolsep}{4pt} 
    \small 
    \begin{tabular}{l|cccccccc}
        \hline
        Metric & \multicolumn{4}{c}{Fidelity} & \multicolumn{4}{c}{Sparsity} \\
        \cmidrule(lr){1-1}\cmidrule(lr){2-5} \cmidrule(lr){6-9}
        Dataset& BA-Shapes & BA-Community & Tree-Cycles & Tree-Grid & BA-Shapes & BA-Community & Tree-Cycles & Tree-Grid \\ \cmidrule(lr){1-1}\cmidrule(lr){2-5} \cmidrule(lr){6-9}
        GNNExplainer & 46.27 $\pm$ 0.78 & 48.32 $\pm$ 0.42 & 93.86 $\pm$ 0.24 & 75.50 $\pm$ 0.53 & 50.12 $\pm$ 0.17 & 54.11 $\pm$ 0.64 & 45.07 $\pm$ 0.28 & 54.38 $\pm$ 0.63 \\ \cmidrule(lr){1-1}\cmidrule(lr){2-5} \cmidrule(lr){6-9}
        PGExplainer & 75.28 $\pm$ 0.73 & 61.23 $\pm$ 0.21 & \textbf{99.22 $\pm$ 0.05} & 87.60 $\pm$ 0.34 & 32.38 $\pm$ 0.11 & 53.08 $\pm$ 0.46 & 55.48 $\pm$ 0.20 & 56.01 $\pm$ 0.62 \\ \cmidrule(lr){1-1}\cmidrule(lr){2-5} \cmidrule(lr){6-9}
        PGM-Explainer & 62.71 $\pm$ 0.23 & 59.35 $\pm$ 0.15 & 95.22 $\pm$ 0.13 & 81.23 $\pm$ 0.61 & 39.56 $\pm$ 0.68 & 42.65 $\pm$ 0.59 & 34.14 $\pm$ 0.24 & 54.84 $\pm$ 0.72 \\ \cmidrule(lr){1-1}\cmidrule(lr){2-5} \cmidrule(lr){6-9}
        Gem& 82.72 $\pm$ 0.24& 55.14 $\pm$ 0.31& 96.38 $\pm$ 0.86 & 86.23 $\pm$ 0.15& 51.67 $\pm$ 0.34& 42.11 $\pm$ 0.24& 35.17 $\pm$ 0.55& 42.72 $\pm$ 0.09\\ \cmidrule(lr){1-1}\cmidrule(lr){2-5} \cmidrule(lr){6-9}
        SubgraphX & 64.04 $\pm$ 0.26 & 59.96 $\pm$ 0.12 & 99.18 $\pm$ 0.08 & 87.63 $\pm$ 0.44 & 42.38 $\pm$ 0.46 & 64.94 $\pm$ 0.45 & 15.38 $\pm$ 0.57 & 26.53 $\pm$ 0.42 \\ \cmidrule(lr){1-1}\cmidrule(lr){2-5} \cmidrule(lr){6-9}
        GISExplainer & \textbf{84.84 $\pm$ 0.53} & \textbf{61.51 $\pm$ 0.23} & 99.19 $\pm$ 0.29 & \textbf{89.10 $\pm$ 0.34} & \textbf{85.75 $\pm$ 0.51} & \textbf{90.61 $\pm$ 0.32} & \textbf{71.37 $\pm$ 0.55} & \textbf{78.24 $\pm$ 0.36} \\ \hline

    \end{tabular}
\end{table*}

\begin{table*}[!tp]
    \centering
    \label{gin_syn}
    \caption{Fidelity and sparsity scores for the top 20 nodes in the test set of synthetic datasets for 2-hop GIN explanations. Higher values signify better performance, with bold black denoting the optimal outcome.}
    \label{tab:GIN_scores}
    \setlength{\tabcolsep}{4pt} 
    \small 
    \begin{tabular}{l|cccccccc}
        \hline
        Metric & \multicolumn{4}{c}{Fidelity} & \multicolumn{4}{c}{Sparsity} \\
        \cmidrule(lr){1-1}\cmidrule(lr){2-5} \cmidrule(lr){6-9}
        Dataset& BA-Shapes & BA-Community & Tree-Cycles & Tree-Grid & BA-Shapes & BA-Community & Tree-Cycles & Tree-Grid \\
        \cmidrule(lr){1-1}\cmidrule(lr){2-5} \cmidrule(lr){6-9}
        GNNExplainer & 71.27 $\pm$ 0.32 & 63.32 $\pm$ 0.42 & 92.22 $\pm$ 0.18 & 87.50 $\pm$ 0.53 & 56.10 $\pm$ 0.32 & 53.11 $\pm$ 0.13 & 53.74 $\pm$ 0.35 & 52.18 $\pm$ 0.51 \\ \cmidrule(lr){1-1}\cmidrule(lr){2-5} \cmidrule(lr){6-9}
        PGExplainer & 81.28 $\pm$ 0.35 & 78.63 $\pm$ 0.92 & 93.22 $\pm$ 0.78 & 87.60 $\pm$ 0.84 & 32.38 $\pm$ 0.11 & 50.28 $\pm$ 0.53 & 48.32 $\pm$ 0.26 & 53.23 $\pm$ 0.24 \\ \cmidrule(lr){1-1}\cmidrule(lr){2-5} \cmidrule(lr){6-9}
        PGM-Explainer & 72.87 $\pm$ 0.27 & 64.89 $\pm$ 0.24 & 96.38 $\pm$ 0.57& 86.57 $\pm$ 0.67& 52.10 $\pm$ 0.62 & 42.11 $\pm$ 0.56 & 45.65 $\pm$ 0.24 & 61.38 $\pm$ 0.56 \\ \cmidrule(lr){1-1}\cmidrule(lr){2-5} \cmidrule(lr){6-9}
        Gem& 81.36 $\pm$ 0.29& 75.42 $\pm$ 0.34& 93.14 $\pm$ 0.81& 84.58 $\pm$ 0.04& 61.41 $\pm$ 0.72& 39.37 $\pm$ 0.38& 47.27 $\pm$ 0.67& 50.48$\pm$ 0.41\\ \cmidrule(lr){1-1}\cmidrule(lr){2-5} \cmidrule(lr){6-9}
        SubgraphX & 74.24 $\pm$ 0.32 & 64.23 $\pm$ 0.24 & 94.18 $\pm$ 0.99 & \textbf{89.32 $\pm$ 0.18} & 41.14 $\pm$ 0.51 & 52.42 $\pm$ 0.46 & 54.68 $\pm$ 0.83 & 38.53 $\pm$ 0.64 \\ \cmidrule(lr){1-1}\cmidrule(lr){2-5} \cmidrule(lr){6-9}
        GISExplainer & \textbf{89.24 $\pm$ 1.35} & \textbf{86.63 $\pm$ 0.87} & \textbf{97.63 $\pm$ 0.63} & 87.63 $\pm$ 0.62 & \textbf{89.75 $\pm$ 0.36} & \textbf{63.53 $\pm$ 0.26} & \textbf{81.46 $\pm$ 0.64} & \textbf{84.23 $\pm$ 1.12} \\ \hline

    \end{tabular}
\end{table*}

Table 2 and 3 provide a detailed comparison of the fidelity and sparsity metrics for various explanation methods applied across different datasets. An optimal explanation method should ideally balance high fidelity, reflecting the method's accuracy, with high sparsity, indicating the method's ability to provide concise explanations. Specifically, Table 2 demonstrates that in the explainability of 2-layer GCN, except for achieving comparable performance with PGExplainer in terms of the evaluation metric fidelity on the Tree-Cycles dataset, GISExplainer has achieved optimal results in terms of both fidelity and sparsity across various datasets. Table 3 presents the experimental results of GISExplainer when applied to 2-layer GIN models. except for achieving a lower fidelity than SubgraphX on the Tree-Grid dataset, the method GISExplainer proposed in this paper has achieved optimal results in terms of both fidelity and sparsity across various datasets. Thus, the consistent superiority in providing accurate and concise explanations for GNNs demonstrates the good performance of our method GISExplainer.


\subsubsection{Experiments on Real-world Datasets}

Table 4 and Table 5 present the fidelity and sparsity values of the explanation methods on real-world datasets. The experimental results show that our method, GISExplainer, has achieved the overall best results, followed by the SubgraphX method which obtained the second-best results. GISExplainer demonstrates the ability to maintain higher fidelity and sparsity on real-world datasets compared to other models, indicating its effectiveness in providing explanations on real-world datasets. It is worth noting that Table 4 and Table 5 exhibit relatively low fidelity scores. This is because, during the model training phase, we only considered the structural information in the data, which resulted in a lower accuracy of the target model and consequently led to an overall decrease in the fidelity values.


To further validate the effectiveness of our model GISExplainer, we conducted experiments under different sparsity levels, and Fig. 6 shows the results. According to the results, regardless of whether the target model is GCN or GIN, our method GISExplainer demonstrated better performance than the comparison methods, indicating that our method can capture faithful explanatory subgraphs that significantly affect the prediction results effectively. Moreover, we observed that our model GISExplainer outperforms other methods even at high sparsity levels, suggesting that GISExplainer can identify significant interaction substructures when requiring the explanatory subgraph to be small in size.


\begin{table*}[htbp]
    \centering
    \caption{Fidelity and Sparsity scores for the top 20 nodes in the test set of real-world datasets for 2-hops GCN explanations. Higher values signify better performance, with bold black denoting the optimal outcome. }
    \label{gcn_real}
    \setlength{\tabcolsep}{6pt}
    \begin{tabular}{l|cccccccc}
        \hline
        Metric & \multicolumn{3}{c}{Fidelity} & \multicolumn{3}{c}{Sparsity} \\
        \cmidrule(lr){1-1}\cmidrule(lr){2-4} \cmidrule(lr){5-7}
        Dataset & Cora & CiteSeer & PubMed & Cora & CiteSeer & PubMed \\
        \cmidrule(lr){1-1}\cmidrule(lr){2-4} \cmidrule(lr){5-7}
        GNNExplainer & 5.98 $\pm$ 0.45 & 8.35 $\pm$ 0.51 & 7.53 $\pm$ 0.73 & 50.28 $\pm$ 0.31 & 53.20 $\pm$ 0.34 & 52.38 $\pm$ 0.20 \\ \cmidrule(lr){1-1}\cmidrule(lr){2-4} \cmidrule(lr){5-7}
        PGExplainer & 6.81 $\pm$ 0.35 & 11.63 $\pm$ 0.92 & 7.22 $\pm$ 0.78 & 42.18 $\pm$ 0.21 & 51.28 $\pm$ 0.53 & 62.32 $\pm$ 0.57 \\ \cmidrule(lr){1-1}\cmidrule(lr){2-4} \cmidrule(lr){5-7}
        PGM-Explainer & 12.23 $\pm$ 0.43 & 13.53 $\pm$ 0.63 & 10.53 $\pm$ 0.31 & 26.12 $\pm$ 0.52 & 43.51 $\pm$ 0.52 & 49.34 $\pm$ 0.25 \\ \cmidrule(lr){1-1}\cmidrule(lr){2-4} \cmidrule(lr){5-7}
        Gem & 10.75 $\pm$ 0.15 & 11.08 $\pm$ 0.57 & 9.72 $\pm$ 0.33 & 62.24 $\pm$ 0.31 & 32.43 $\pm$ 0.09 & 56.24 $\pm$ 0.42 \\ \cmidrule(lr){1-1}\cmidrule(lr){2-4} \cmidrule(lr){5-7}
        SubgraphX & 9.24 $\pm$ 0.24 & 10.23 $\pm$ 0.19 & 7.58 $\pm$ 0.37 & \textbf{90.52 $\pm$ 0.11} & 84.53 $\pm$ 0.46 & \textbf{91.68 $\pm$ 0.41} \\ \cmidrule(lr){1-1}\cmidrule(lr){2-4} \cmidrule(lr){5-7}
        GISExplainer & \textbf{18.21 $\pm$ 0.28} & \textbf{15.12 $\pm$ 0.57} & \textbf{19.63 $\pm$ 0.38} & 84.75 $\pm$ 0.57 & \textbf{86.57 $\pm$ 0.18} & 81.46 $\pm$ 0.64 \\ \hline

    \end{tabular}
\end{table*}

\begin{table*}[htbp]
    \centering
    \label{gin_real}
    \caption{Fidelity and Sparsity scores for the top 20 nodes in the test set of real-world datasets for 2-hop GIN explanations. Higher values signify better performance, with bold black denoting the optimal outcome. }
    \setlength{\tabcolsep}{6pt}
    \begin{tabular}{l|cccccc}
        \hline
        Metric & \multicolumn{3}{c}{Fidelity} & \multicolumn{3}{c}{Sparsity} \\
        \cmidrule(lr){1-1}\cmidrule(lr){2-4} \cmidrule(lr){5-7}
        Dataset & Cora & CiteSeer & PubMed & Cora & CiteSeer & PubMed \\
        \cmidrule(lr){1-1}\cmidrule(lr){2-4} \cmidrule(lr){5-7}
        GNNExplainer & 4.98 $\pm$ 0.74 & 7.67 $\pm$ 0.82 & 5.53 $\pm$ 0.37 & 50.24 $\pm$ 0.27 & 57.72 $\pm$ 0.27 & 52.37 $\pm$ 0.89 \\ \cmidrule(lr){1-1}\cmidrule(lr){2-4} \cmidrule(lr){5-7}
        PGExplainer & 8.23 $\pm$ 0.41 & 10.21 $\pm$ 0.74 & 5.22 $\pm$ 0.51 & 32.38 $\pm$ 0.11 & 50.28 $\pm$ 0.53 & 48.32 $\pm$ 0.26 \\ \cmidrule(lr){1-1}\cmidrule(lr){2-4} \cmidrule(lr){5-7}
        PGM-Explainer & 10.21 $\pm$ 0.45 & 11.83 $\pm$ 0.33 & 9.15 $\pm$ 0.27 & 62.15 $\pm$ 0.18 & 52.51 $\pm$ 0.52 & 58.34 $\pm$ 0.25 \\ \cmidrule(lr){1-1}\cmidrule(lr){2-4} \cmidrule(lr){5-7}
        Gem & 6.63 $\pm$ 0.42& 9.46$\pm$ 0.32& 14.24 $\pm$ 0.61 & 53.13 $\pm$ 0.95 & 61.74 $\pm$ 0.25 & 61.35 $\pm$ 0.61 \\ \cmidrule(lr){1-1}\cmidrule(lr){2-4} \cmidrule(lr){5-7}
        SubgraphX & 8.61 $\pm$ 0.14 & 11.45 $\pm$ 0.25 & 8.58 $\pm$ 0.21 & 82.14 $\pm$ 0.11 & 86.21 $\pm$ 0.57 & \textbf{90.15 $\pm$ 0.83} \\ \cmidrule(lr){1-1}\cmidrule(lr){2-4} \cmidrule(lr){5-7}
        GISExplainer & \textbf{17.21 $\pm$ 0.16} & \textbf{13.12 $\pm$ 0.27} & \textbf{17.63 $\pm$ 0.27} & \textbf{85.27 $\pm$ 0.36} & \textbf{87.53 $\pm$ 0.26} & 81.46 $\pm$ 0.64 \\ \hline

    \end{tabular}
\end{table*}

\begin{figure*}[!tbp]
	\begin{flushleft}
		\centering
            \label{vis_2}
		\includegraphics[width=1\linewidth]{Fig6-fidelity.jpg}
		\caption{The fidelity scores, derived from the application of the GCN and GIN model on various real-world datasets.}
		\label{control}
	\end{flushleft}
\end{figure*}

\begin{figure*}[htbp]\centering
	\centering
	\label{adv_analysis}
	{ \includegraphics[width = 7.0in]{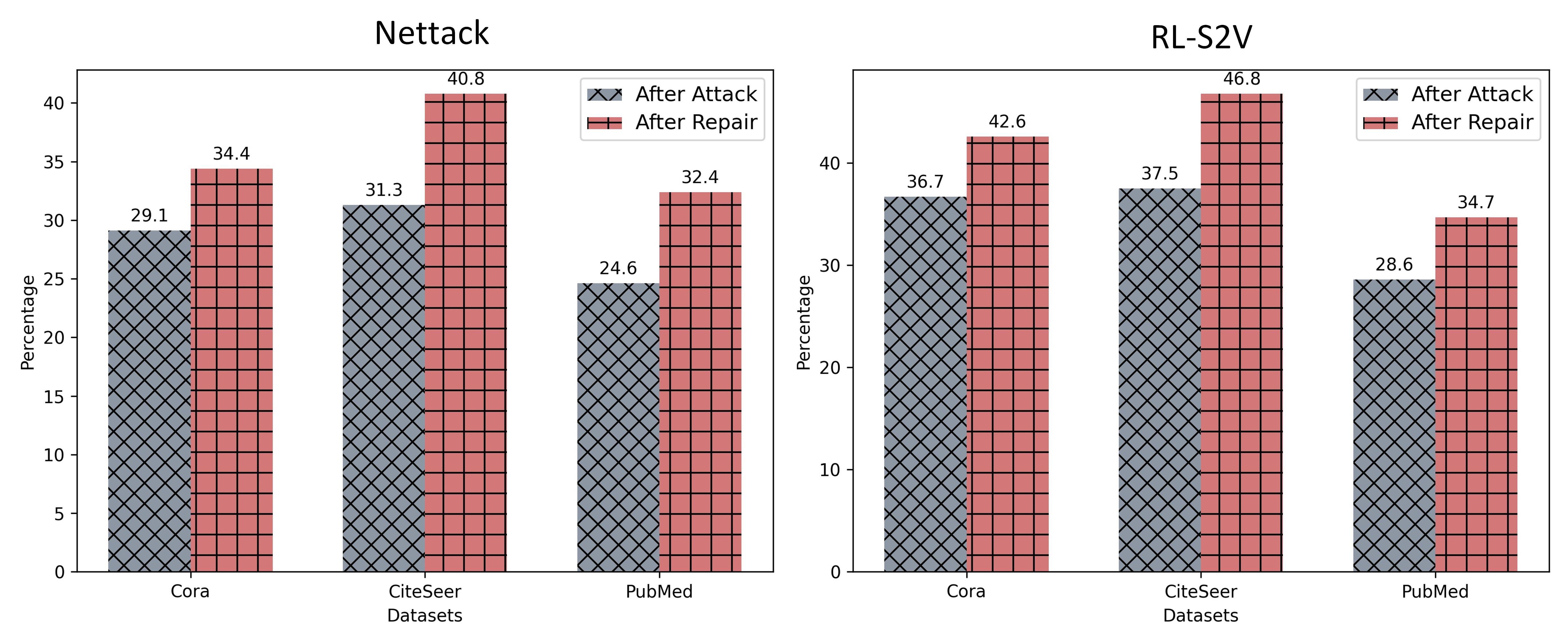}}
	\caption{The effectiveness of model explainability for adversarial attacks.}
	\label{framework}
\end{figure*}

\begin{table*}[!ht]
	\caption{The average time taken to explain the top 5 nodes in the test set (In Second)}
        \label{time_compute}
	\centering
	\begin{tabular}{@{}lcccccccc@{}}
		\hline
		{Models}& {BA-Shapes} & {BA-community} & {Tree-cycle} & {Tree-grid} & {Cora} & {CiteSeer} & {PubMed}\\
		\hline
		\ GISExplainer without optimization   & 4349 & 5149 & 1202 & 2103 & 48242 & OOM & OOM \\ \hline
		\ GISExplainer   & 502  & 519 & 51 & 135 & 3154 & 3527 & 6420 \\
		\hline
	\end{tabular}
	\label{tab:datasets}
\end{table*}

\subsection{Model Explainability for Adversarial Attacks}
On clean graphs, GNNs explanation methods can produce explanatory subgraphs for specific instances that contribute to the model's output. However, in adversarial attack scenarios, GNNs may produce unreliable predictions due to adversarial perturbations in the input graphs. In such cases, the model explanation methods can be used to identify explanatory subgraphs that lead to the model's unreliable predictions, thereby discovering the adversarial perturbations and enabling adversarial defenses against them. Therefore, we conducted experiments on explanatory subgraph discovery and adversarial perturbation tracing based on the explanation methods. Here, two representative adversarial attack methods Nettack \cite{nettack} and RL-S2V \cite{rl2sv} were adopted, and 10 nodes were selected from the test sets of Cora, CiteSeer, and PubMed as the target nodes for both attack and explanation. Specifically, we utilized the GNNs explanation methods to generate explanatory subgraphs for the unreliable predictions under targeted adversarial attacks.
Then we regard the explanatory subgraphs as the perturbed substructures and perform restoration operations on them. Subsequently, we input the restored graphs into the target GNNs and compare the difference in prediction accuracy caused by them compared to the original perturbed graphs.

The experimental results are shown in Fig. 7. In this paper, for the obtained explanatory subgraphs regarding adversarial attacks, a unified method based on local smoothness is adopted to remove the perturbations. The results demonstrate that the identification and defense of adversarial perturbations based on explanatory subgraphs effectively improve the performance, i.e., classification accuracy, of the target model. This illustrates the credibility of the explanatory subgraphs and thereby proves the effectiveness and applicability of the proposed explanation method GISExplainer for explaining and defending against adversarial attacks on GNNs.

\subsection{\label{sec:level1} Complexity and Time Analysis}
Here, we analyze the computational complexity of the proposed method. GISExplainer primarily consists of three parts: causal screening mechanisms, game-theoretic interaction calculation mechanism, and heuristic search strategy. Generally, GISExplainer determines candidate edges at each step based on the heuristic search strategy, and then selects salient edges for updating the explanatory subgraph during the causal Screening process based on the game-theoretic interaction based subgraph importance. Therefore, the time complexity primarily focuses on determining candidate edges and calculating the importance of subgraphs. Specifically, for the $l$-hop subgraph of a target node $v$ with $M$ edges, the maximum number of candidate edges is equivalent to $M$. Additionally, for each candidate edge's corresponding candidate explanatory subgraph, the Shapley value of its contained coalition needs to be calculated to obtain the importance of the subgraph. Thus, the complexity depends on the number of possible combinations consisting of more than one variable that does not exceed $2^Q$, where $Q$ is the number of edges in candidate explanatory subgraph. Thus, the upper limit of time complexity of GISExplainer is $O(M*2^Q)$. Due to the connectivity constraints mentioned in this paper, as shown in Equations (13) and (14), the actual number is much smaller than this. In addition, we count the time consumption of GISExplainer with and without the efficiency optimization mechanism in Table 6. Compared to the method without using the efficiency optimization mechanism, we can find that the efficiency of GISExplainer has been significantly improved, i.e., save approximately $90\%$ of the time, after applying the efficiency optimization. 

\section{Conclusion}
To address the problem of explainability in GNNs, a novel game-theoretic interaction-based explanation method, namely GISExplainer, is designed in this paper so as to extract faithful explanatory subgraphs effectively. Unlike existing perturbation-based GNNs explanation methods, GISExplainer proposes a subgraph importance measurement method based on the intensity of game interaction within coalitions, which integrates both positive and negative effects. This method is further utilized for salient edges selection and explanatory subgraph construction successively. To construct the explanatory subgraph, a heuristic graph search algorithm is introduced. Extensive experimental results demonstrate the superiority of the proposed method over state-of-the-art models and the applicability of the method for explaining adversarial attacks.

Our future work mainly includes the following several aspects. First, similar to other explanation methods based on Shapley values, the computational cost of GISExplainer poses a challenge, and it is necessary to design more efficient explanation methods in the future. Second, the robustness and generalization of GNNs are hot topics in the research field, and how to conduct explanatory analysis of robustness and generalization is a question worthy of future study.

\ifCLASSOPTIONcaptionsoff
\fi

\bibliographystyle{IEEEtran}
\bibliography{document}


\vspace{-40pt}
\begin{IEEEbiography}[{\includegraphics[width=1in,height=1.25in,clip,keepaspectratio]{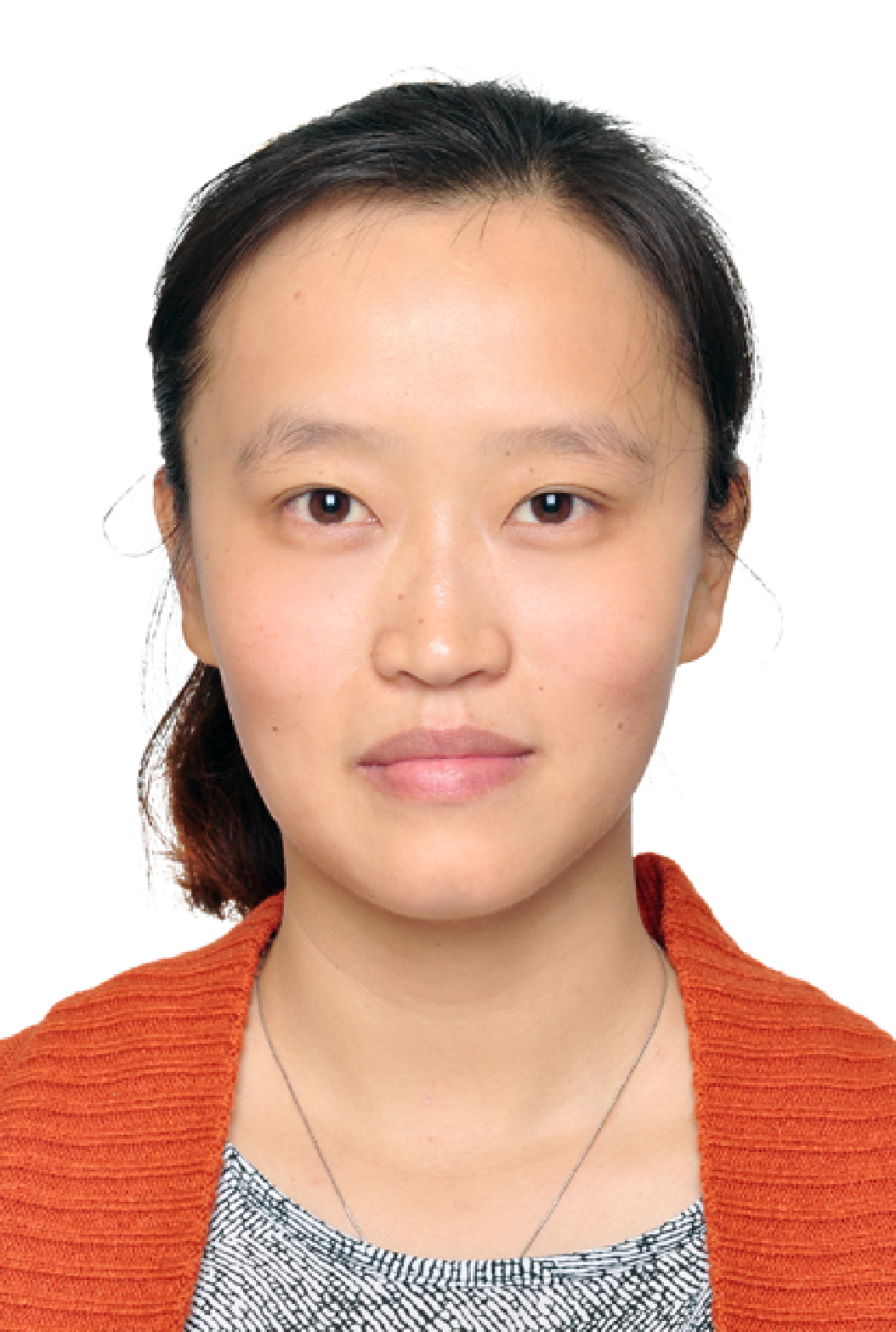}}]
{Xingping Xian} received her M.S. degree from Chongqing University of Posts and Telecommunications, Chongqing, China and is currently a Ph.D. candidate of the school of computer science, Chongqing University of Posts and Telecommunications. Her current research interests include machine learning, social network analysis, privacy-preserving, etc. She has published more than ten scientific papers in international journals and conferences. 
\end{IEEEbiography}

\vspace{-40pt}
\begin{IEEEbiography}[{\includegraphics[width=1in,height=1.25in,clip,keepaspectratio]{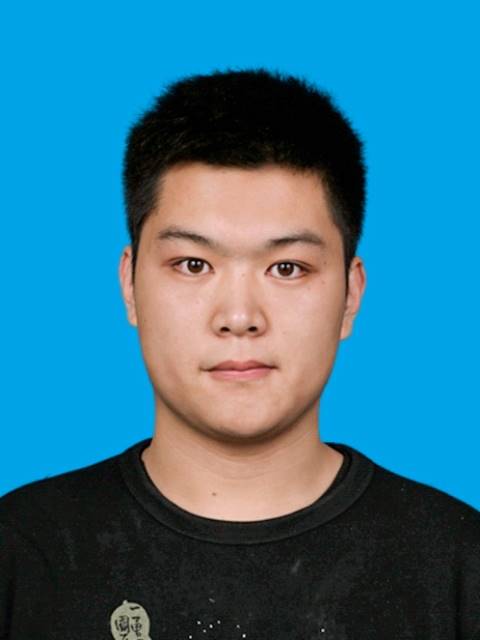}}]
{Jianlu Liu } received the bachelor's degree from Linyi University, Linyi, China, in 2022. He is currently pursuing the master's degree with the School of Cyber Security and Information Law in Chongging University of Posts and Telecommuni cations, Chongqing, China. His research interests include graph neural networks and graph neural network explainability.
\end{IEEEbiography}

\vspace{-40 pt}
\begin{IEEEbiography}[{\includegraphics[width=1in,height=1.25in,clip,keepaspectratio]{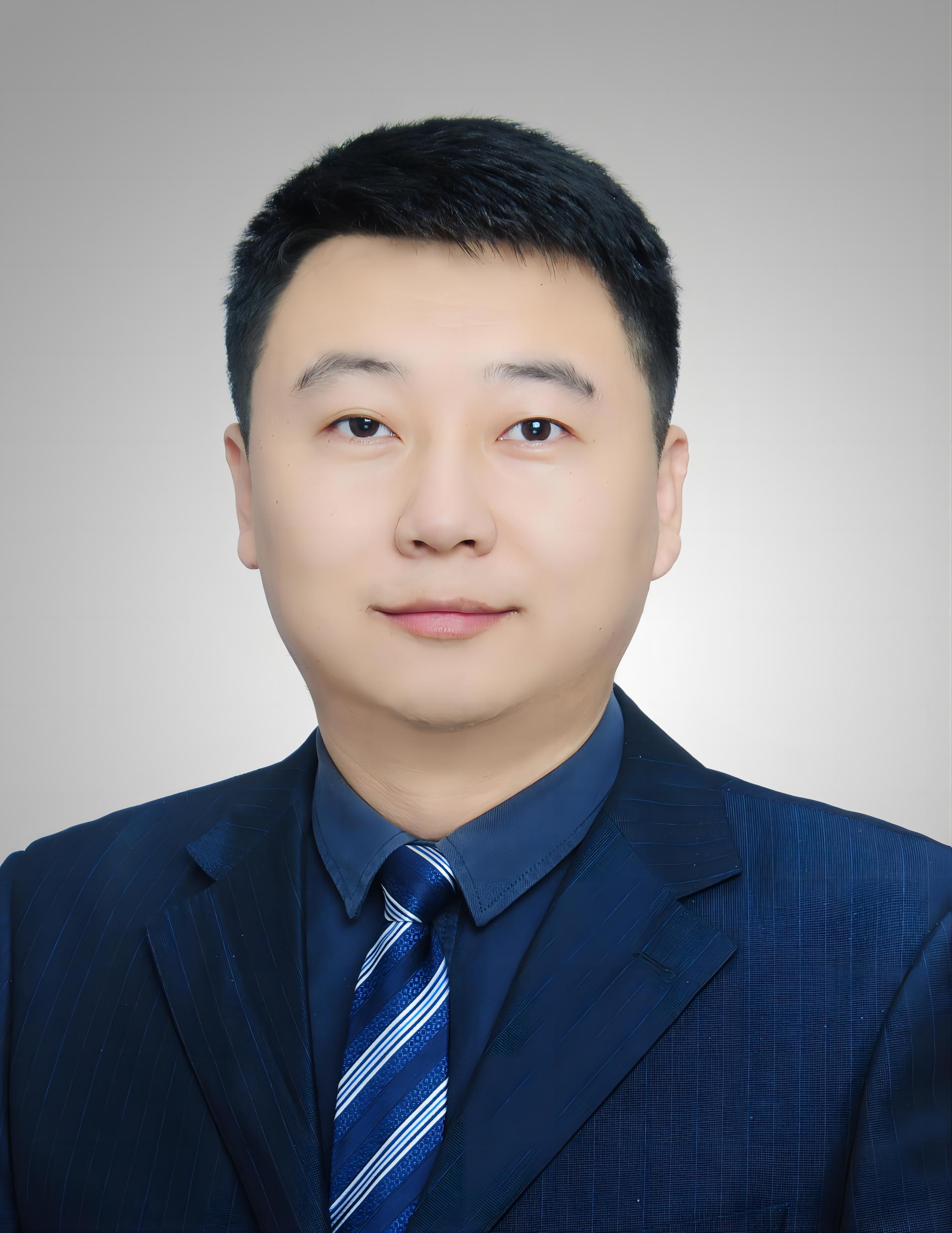}}]
{Tao Wu} received the Ph.D. degree from University of Electronic Science and Technology of China, in June 2017. He is currently the Head and a Professor at Department of Cybersecurity, School of Cyber Security and Information Law, Chongqing University of Posts and Telecommunications, China. He has authored more than 50 high quality papers, including IEEE Transactions on IE, PR, etc. His research interests include graph neural networks, artificial intelligence (AI) security, graph mining. 
\end{IEEEbiography}

\vspace{-40pt}
\begin{IEEEbiography}[{\includegraphics[width=1in,height=1.25in,clip,keepaspectratio]{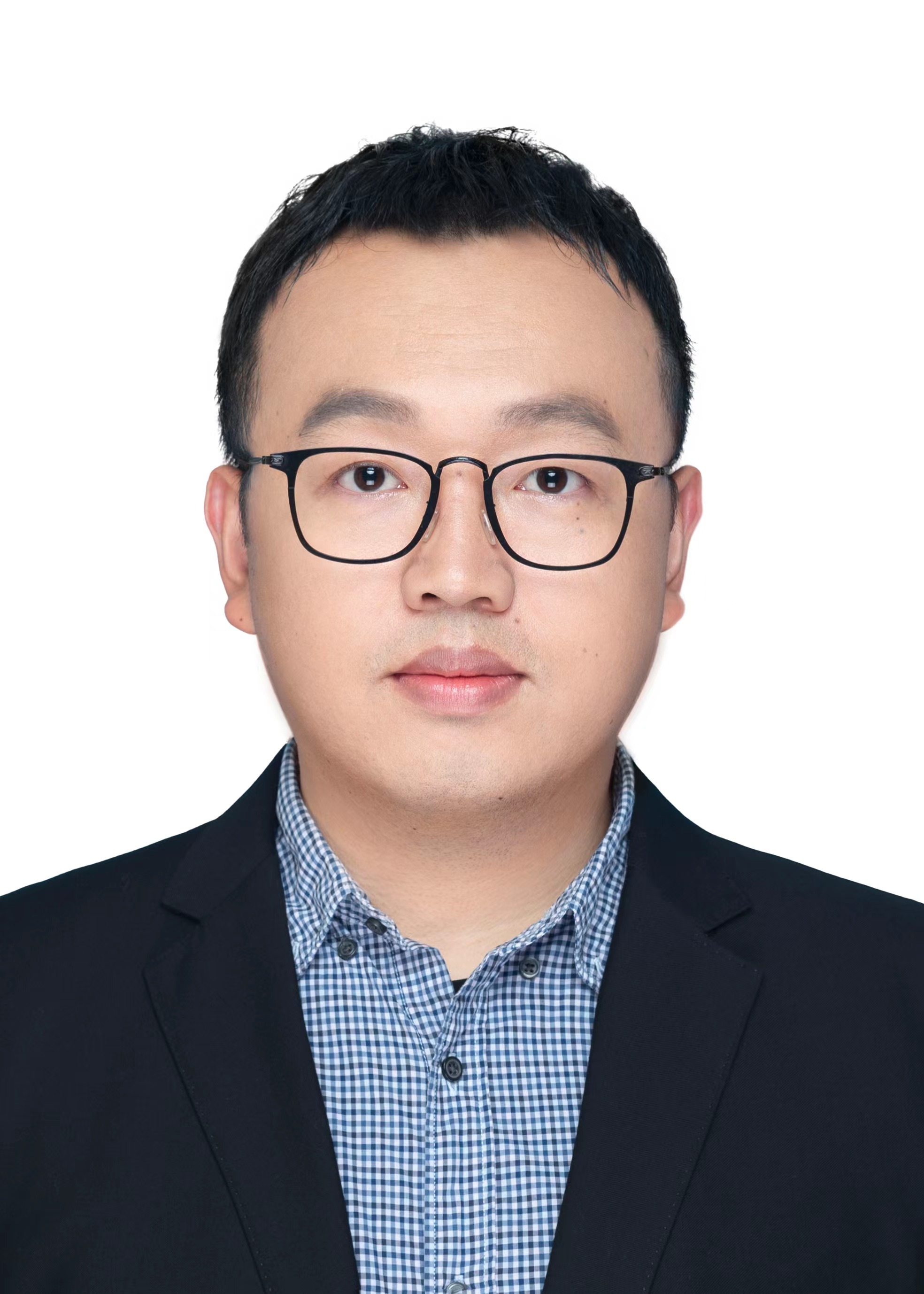}}]
{Chao Wang} received the Ph.D. degree in computer science from the University of Electronic Science and Technology of China, Chengdu, China, in 2019. In 2019, he joined the School of Electrical Engineering, Chongqing University, as an Assistant Professor. He is currently with the School of Computer and Information Science, Chongqing Normal University, Chongqing, China. His research interests include data mining and machine learning and its applications.
\end{IEEEbiography}

\vspace{-40pt}
\begin{IEEEbiography}[{\includegraphics[width=1in,height=1.25in,clip,keepaspectratio]{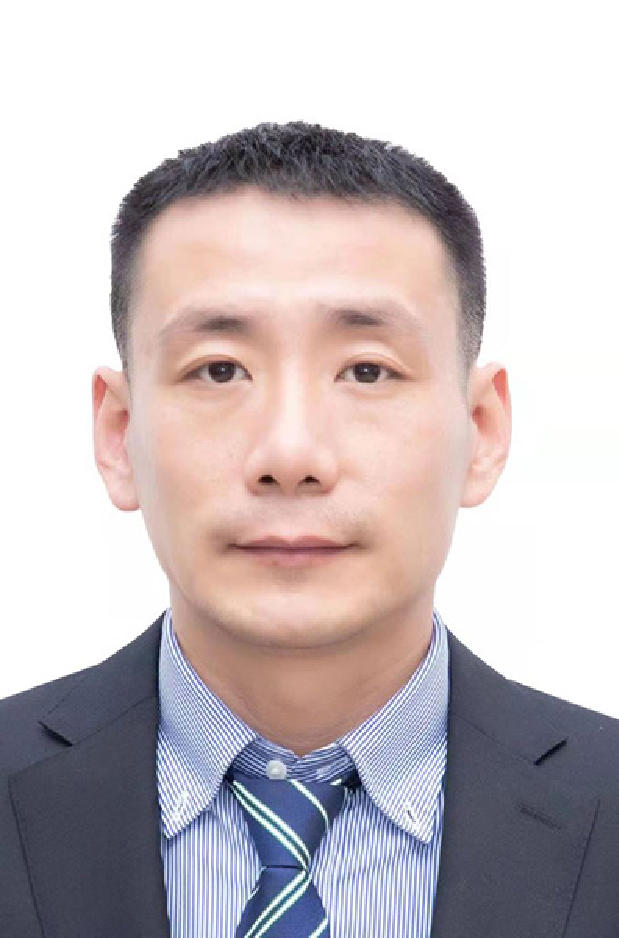}}]
{Shaojie Qiao} received the B.S. and Ph.D. degrees from Sichuan University, Chengdu, China, in 2004 and 2009, respectively. He is a Distinguished Young Scholars of Sichuan Province. He is currently a professor with the School of Software Engineering, Chengdu University of Information Technology, Chengdu, China. He has authored more than 160 high quality papers, including IEEE Transactions on ITS, TKDE, TNNLS, etc. His research interests include graph neural networks and spatio-temporal databases.
\end{IEEEbiography}

\vspace{-40pt}
\begin{IEEEbiography}[{\includegraphics[width=1in,height=1.20in,clip,keepaspectratio]{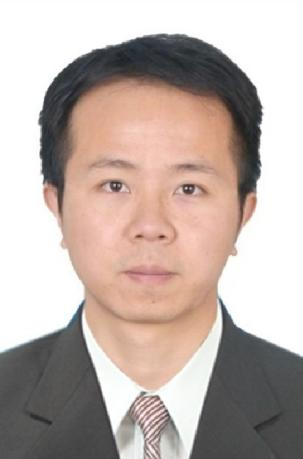}}]
{Xiaochuan Tang} received the Ph.D. degree from the University of Electronic Science and Technology of China, Chengdu, China. He used to be a visiting professor at University of Padova, Italy. He is currently an Associate Professor at Chengdu University of Technology, Chengdu, China. He has published more than 30 scientific articles in international journals and conferences. His research interests include deep learning, remote sensing and geohazard prevention.
\end{IEEEbiography}

\vspace{-40pt}
\begin{IEEEbiography}[{\includegraphics[width=1in,height=1.20in,clip,keepaspectratio]{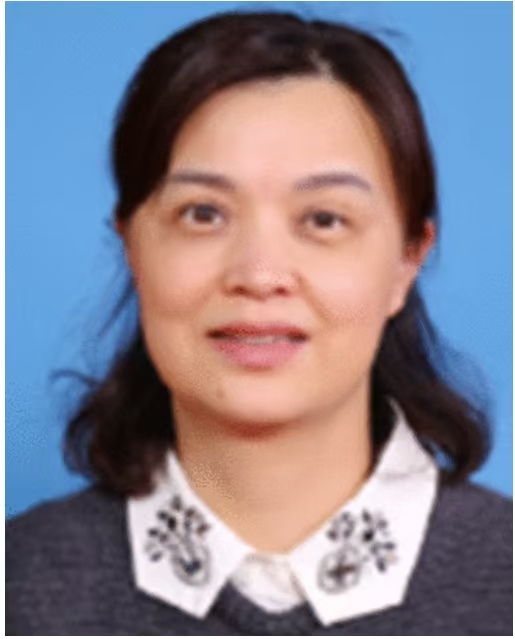}}]
{Qun Liu} received the B.S. degree from Xian Jiaotong University, Xian, China, in 1991, and the M.S. degree from Wuhan University, Wuhan, China, in 2002, and the Ph.D. degree from Chongqing University, Chongqing, China, in 2008. She is currently a Professor with Chongqing University of Posts and Telecommunications, Chongqing. Her current research interests include complex and intelligent systems, neural networks, and intelligent information processing.
\end{IEEEbiography}

\end{document}